\pdfoutput=1
\documentclass[11pt]{article}

\usepackage[]{authblk}
\usepackage{mdframed}
\usepackage{setspace}
\usepackage{multirow}
\usepackage[table,xcdraw]{xcolor}

\makeatletter
\renewcommand\AB@affilsepx{, \protect\Affilfont}
\makeatother

\usepackage{tikz}

\usepackage[]{ACL2023}

\usepackage{times}
\usepackage{latexsym}
\usepackage{graphicx}
\usepackage{subcaption}

\usepackage{tabu}
\usepackage{multirow}
\usepackage{makecell} 

\usepackage[T1]{fontenc}
\usepackage[utf8]{inputenc}

\usepackage{microtype}

\usepackage{inconsolata}
\usepackage{csquotes}

\usepackage{todonotes}
\usepackage{fdsymbol}

\newcommand{\pz}{\hphantom{0}}
\newcommand{\pzz}{\hphantom{00}}

\title {Where Do We Go from Here? Multi-scale Allocentric Relational Inference \\from Natural Spatial Descriptions}
\author[1a]{Tzuf Paz-Argaman}
\author[b]{Sayali Kulkarni}
\author[b]{John Palowitch}
\author[b]{Jason Baldridge}
\author[a,b]{\authorcr Reut Tsarfaty}
\affil[a]{Bar-Ilan University} 
\affil[b]{Google}
\affil[ ]{\authorcr \tt \{tzuf.paz-argaman, reut.tsarfaty\}@biu.ac.il}
\affil[ ]{\authorcr \tt \{palowitch, sayali, jasonbaldridge\}@google.com}

\date{}

\begin{document}

\maketitle

\setcounter{footnote}{1}

\begin{abstract}

When communicating routes in natural language, the concept of {\em acquired spatial knowledge} is crucial for geographic information retrieval (GIR) and in spatial cognitive research.
However, NLP navigation studies often overlook the impact of such acquired knowledge on textual descriptions.
Current navigation studies concentrate on  egocentric local descriptions (e.g., `it will be on your right') that require reasoning over the agent's local perception. These instructions are typically given as a sequence of steps, with each action-step explicitly mentioning and being followed by a landmark that the agent can use to verify they are on the right path (e.g., `turn right and then you will see...').  
In contrast, descriptions based on knowledge acquired through a map provide a complete  view of the environment and capture its overall structure. These instructions  (e.g., `it is south of Central Park and a block north of a police station') are typically  non-sequential, contain allocentric relations, with  multiple spatial relations and implicit actions, without any explicit verification. This paper introduces the Rendezvous (RVS) task and dataset, which includes 10,404 examples of English geospatial instructions for reaching a target location using map-knowledge. Our analysis reveals that RVS exhibits a richer use of spatial allocentric relations, and requires resolving more spatial relations simultaneously compared to previous text-based navigation benchmarks.\footnotetext{This work was done partly during an internship at Google Research.}\footnote{Data and code:  \url{https://github.com/OnlpLab/RVS}.}

\end{abstract}

 \begin{figure}[!t]

  %\begin{subfigure}
  % \centering
  %\scalebox{0.49}{
  %          \includegraphics[width=1 \textwidth]{images/map_1.png}}
  %\end{subfigure}

  \centering
  %\begin{subfigure}
  %A37GOI3N77WX213XD2A6FGFNVI66D9NCSNE6ZCBY09SC1
\scalebox{0.47}{
            \includegraphics[width=1 \textwidth]{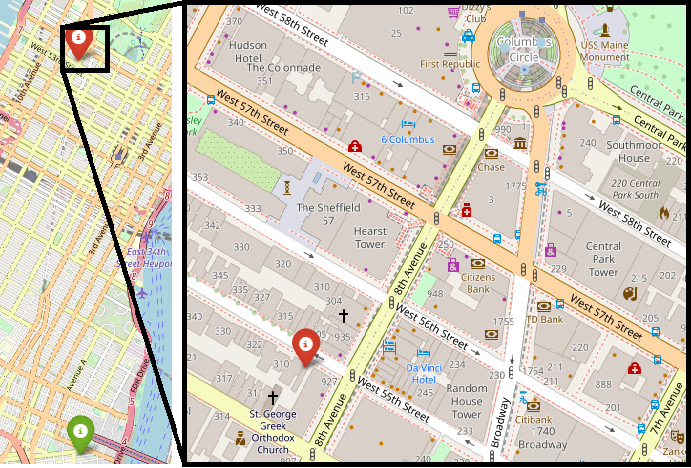}}

              \vspace*{-1mm}
                 {\footnotesize{

                 \begin{spacing}{0.6}
                
                 \begin{flushleft} 
 \begin{mdframed}
 
\textit{\textbf{I'm pretty far away, almost all the way to Central Park, just 3–4 blocks from Columbus Circle. Walk north on 8th Ave., and I'm at a parking entrance a block north of a police station.
 } }
 \end{mdframed}
 \end{flushleft} 
\end{spacing} }}

%                  {\scriptsize{ \begin{spacing}{0.5}
                 
%                  \begin{flushleft} 
% \textit{\textbf{I'm pretty far away, almost all the way to Central Park, just 3-4 blocks from Columbus Circle. Walk north on 8th Ave. and I'm at a parking entrence a block north of a police station.
% } }
%  \end{flushleft}
% \end{spacing} }}
%\end{subfigure}

        \caption
        { 
        % An illustration of the RVS task.
       An illustration example from the RVS dataset. The RVS input consists of (1) a bird's-eye instruction of the goal location (shown at the bottom), (2) a starting point (green marker), and (3) a map representation of the environment. The output is the goal (red marker).}

        \label{fig:main_example}
    \end{figure}
\section{Introduction}

% However, people often arrange rendezvous locations by giving non-sequential descriptions with an incomplete set of actions. These bird's-eye view descriptions of goal geolocations, include references to prominent spatial features and salient landmarks, e.g. `The hotel is five blocks south of the Empire State Building, just across the street from a park.`  Abstract entities such as `block` and spatial features such as cardinal directions `south` are more common in high-level instructions than in low-level sequential instructions.  

In today’s world, cell phones with powerful mapping applications are widely used. However, even with this technology at our fingertips, many people still rely on geospatial instructions to arrange rendezvous locations by providing natural language descriptions that reference landmarks and their geospatial relation, e.g., `...a block north of a police station' (Figure \ref{fig:main_example}).
Retrieving locations and paths from natural language spatial descriptions is essential for disaster areas \cite{hu2023geo}, for the billions of people without addresses \cite{upu:2012, abebrese2019implementing}, and for Geographic Information Retrieval (GIR), especially from the web \citep{spink2002sex, sanderson2004analyzing}.  
%which struggles to retrieve spatial information from unstructured texts. 

% especially from the web were an estimation of ~20\% web queries contain geospatial terms \cite{sanderson2004analyzing, spink2002sex}.
% \par

% Before the wide availability of cell phones with powerful mapping applications, people often arranged rendezvous locations by giving descriptions that reference salient landmarks, e.g. “the coffee roaster a block south of the square just west of the Empire State Building.” 
% With billions of people living without an address according to the \citet{upu:2012}, it is no wonder that even with cell phones giving geospatial directions is still a common practice. 

In spatial cognitive research, it is widely accepted that spatial language is associated with cognitive representations of the environment and originates from spatial memory \cite{hayward1995spatial}. Thus, navigation instructions are affected by the way individuals acquire spatial knowledge over their environment 
%, that is, the acquired knowledge is encoded in the instructions 
\citep{tversky2005functional, thorndyke1982differences, kuipers1978modeling}. The dominant theory for spatial knowledge acquisition, that of   \citet{siegel1975development}, describes three levels of human knowledge about their environment:
(i)~\textit{Landmark knowledge}: the ability to describe the characteristics of distinct objects, which may be located along a route, without indicating the relationship or path between those landmarks, (ii)~\textit{Route knowledge}: includes sequential information such as  directions for navigation instructions,
% from one location to another. 
and finally (iii)~{\em Survey knowledge}, which involves understanding the layout and composition of the environment and describing landmarks in relation to one another using an external reference system, such as the directional relationships between landmarks.

Instructions based on {\em survey knowledge} contain a bird's-eye view perception of the environment. These higher-level descriptions involve allocentric relations and cardinal directions (`east of'), are non-sequential, with implicit actions and multiple spatial relations without any verification (e.g.,  `3–4 blocks north of Columbus Circle and north of a police station'). They  require geospatial numerical reasoning (`two buildings from') and understanding of complex shapes such as `Y-shaped street' \cite{jayannavar2020learning, lachmy2022draw}.
They contain a mix of indefinite descriptions referencing  salient landmarks (`a building'), as well as 
proper names (`the Empire State Building').

Despite the importance of geospatial instructions in daily life, current NLP  geospatial datasets lack instructions that encompass  all such levels of acquired knowledge \cite{chen2019touchdown}.
% of often lack complex geospatial instructions that require survey knowledge and geospatial compositional reasoning. 
While many NLP geolocation tasks primarily involve instructions based on {\em landmark knowledge} \cite{wing2014hierarchical}, 
% which is the first level and includes descriptions of the destination's properties and its immediate surroundings.
text-based navigation tasks focus on the second level --- {\em route knowledge} --- with step-by-step local perception \cite{ku2020room}. However, current spatial datasets are missing the third level –- {\em survey knowledge} --- which involves global perception and requires reasoning over multiple spatial relations simultaneously.

% descriptions including configurational properties of the environment are scarce in current geospatial datasets in NLP. Accordingly, the ability to handle complex geospatial descriptions based on survey knowledge has not been a major challenge in current datasets. 

% Thus, giving a geospatial description of a rendezvous location based on named
% and vague references to entities and spatial features, is still a common practice. 
% \par 

%with 10,404 instructions

Here, we introduce the {\em Rendezvous} (RVS) task to advance systems that can interpret high-level {\em survey knowledge}-based navigation instructions that require global spatial  reasoning. 
% our goal is to advance systems that can interpret instructions based on survey knowledge.
% %, with non-specific names near the goal location.
% We introduce the {\em Rendezvous} (RVS) task, aimed at interpreting high-level navigation instructions that require spatial allocentric reasoning. 
The input of the task is a starting point, a non-sequential instruction of a rendezvous location, and a map. The goal is to retrieve the coordinates of the rendezvous point.
We crowdsourced 10,404 rendezvous instructions. To gather instructions based on survey knowledge, we presented participants with a map that provided them with precise information that would have otherwise required extensive exploration of the environment \citep{thorndyke1982differences, uttal2000seeing, plumert2007emerging, tversky1996spatial}. 

We collected instructions over three cities in the USA: Manhattan, Pittsburgh and Philadelphia. The use of multiple cities allows for a realistic zero-shot setup where a model is trained on one city and tested on another  city unseen during training. 
This new zero-shot setup is a challenging testbed for models' ability to generalize to new environments. This is also relevant for handling changing environments \cite{zhang2021situatedqa}. It is part of our contribution to create a realistic and challenging setup and show that current models do not suffice in addressing this multifaceted challenge.

Our linguistically-driven analysis shows that the RVS task requires 
significantly more spatial allocentric reasoning,  resolving more spatial relations simultaneously, and with fewer explicit actions and state verifications, compared with previous text-based navigation benchmarks
\citep{paz2019run, chen2019touchdown, ku2020room}. 

\par

\section{The RVS Task and Environment}
\label{sec:task}
In this work we address the task of following geospatial instructions given in colloquial language based on a dense urban map. The input to the RVS task is as follows: (i) a map with rich details, given as a knowledge graph; (ii) an explicit starting point, given in coordinates (latitude and longitude); and (iii) a geospatial instruction describing the location of the goal in relation to the landmarks on the map and the given starting point. The output of the RVS task is the coordinates of the goal within the boundaries of the map.

The map was created using  \href{http://www.openstreetmap.org}{OpenStreetMap (OSM)}.\footnote{OSM is a user-updated map of the world.  \url{http://www.openstreetmap.org}} We extracted landmarks and streets and connected them to form a graph. To connect landmarks that do not intersect with streets, we projected the landmarks onto the nearest streets (up to four) and added the projected nodes and edges connecting the landmarks and projections to the graph.

% We then added the streets as edges where the end and begging of the street are connected nodes. We add a landmark as a node to the graph and then we project the landmark on up to four closest streets, we then add the projected nodes and the edges connecting the landmark and the projection. 

% The task we are considering is one where we have descriptions of goal locations relative to a starting point, and 1 or more landmarks.

\begin{figure*}[!th]
  \centering
  \scalebox{1}{
  \begin{subfigure}[b]{0.33\textwidth}
    \includegraphics[width=\textwidth]{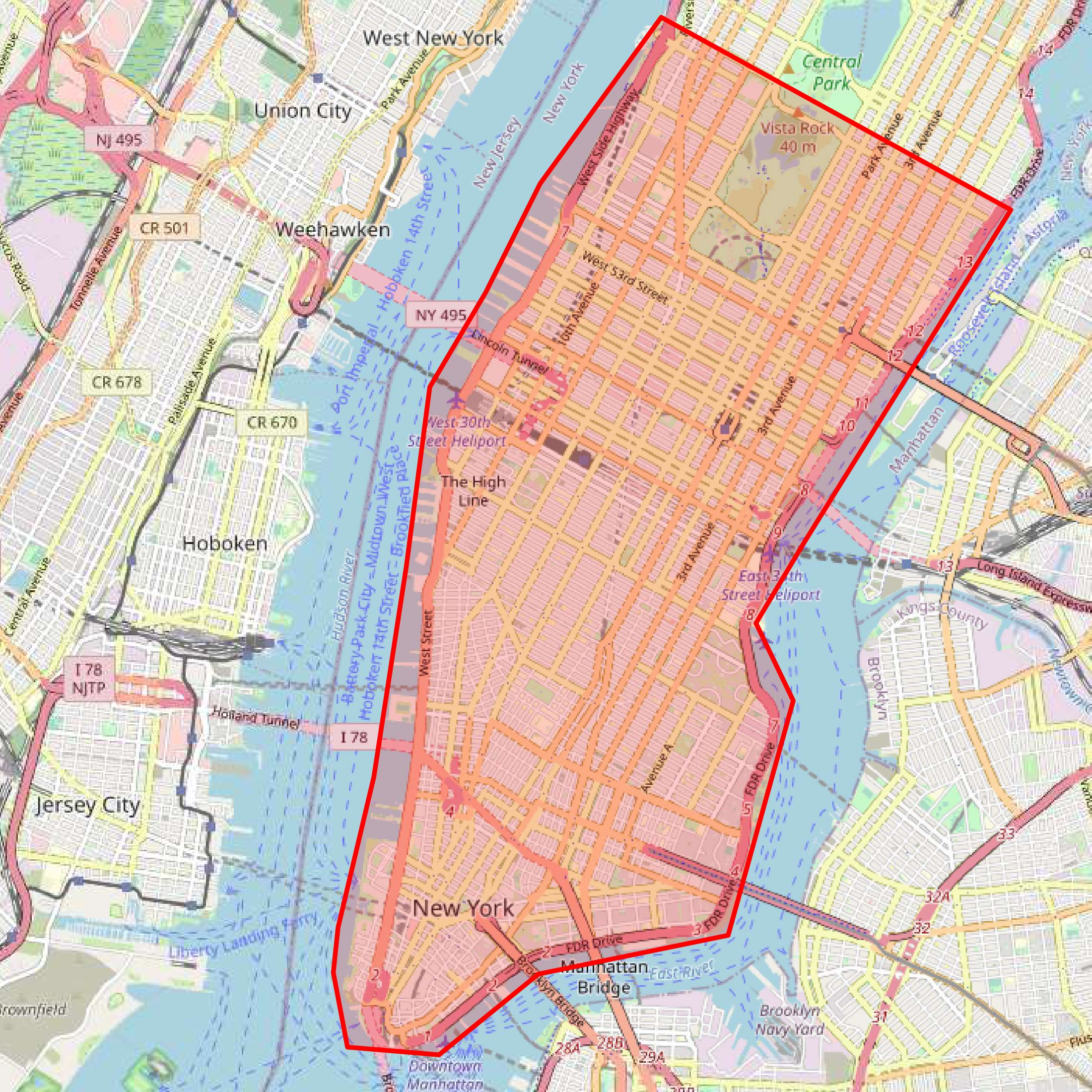}
    \caption{Manhattan}
    \label{fig:image1}
  \end{subfigure}
  \hfill
  \begin{subfigure}[b]{0.33\textwidth}
    \includegraphics[width=\textwidth]{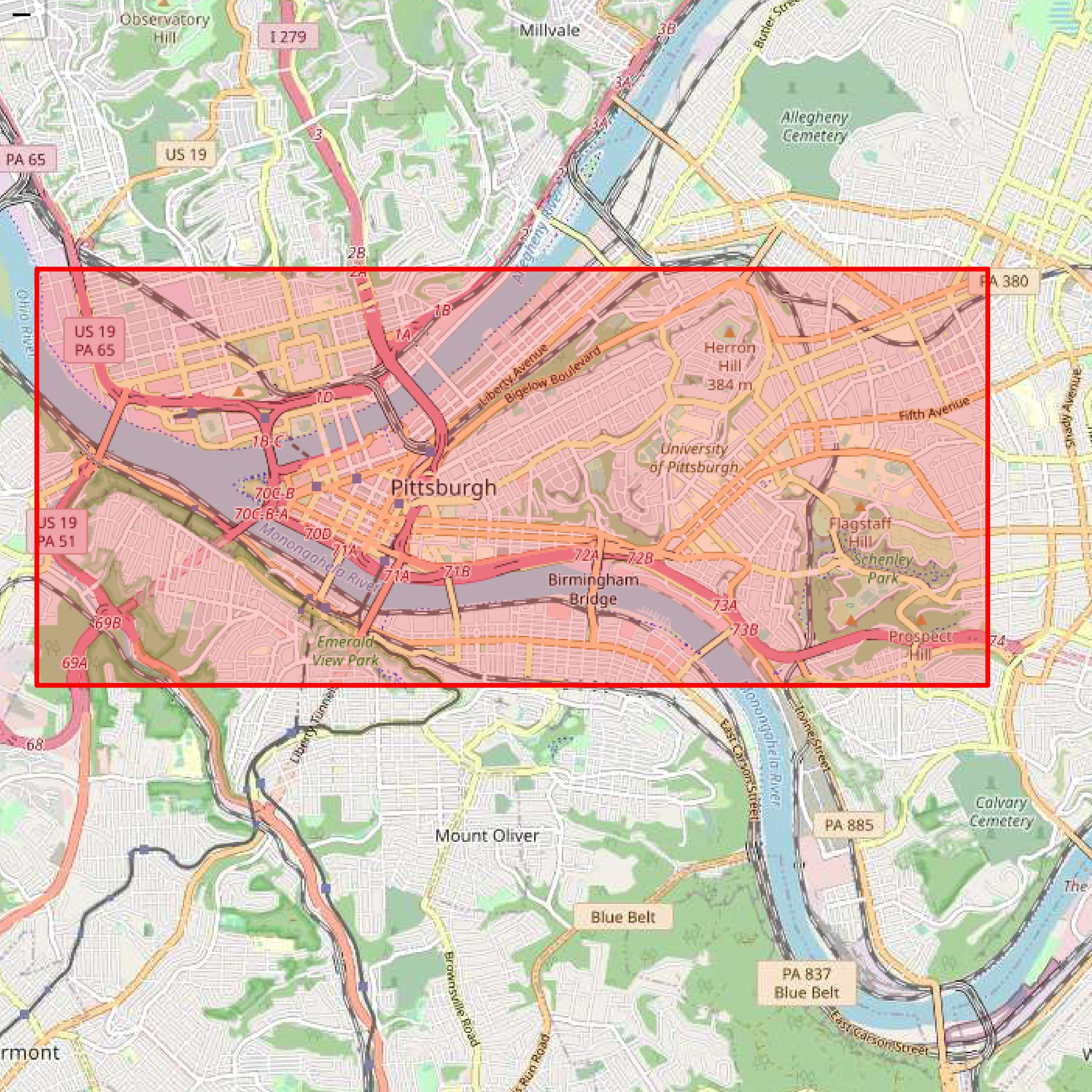}
    \caption{Pittsburgh}
    \label{fig:image2}
  \end{subfigure}
  \hfill
  \begin{subfigure}[b]{0.33\textwidth}
    \includegraphics[width=\textwidth]{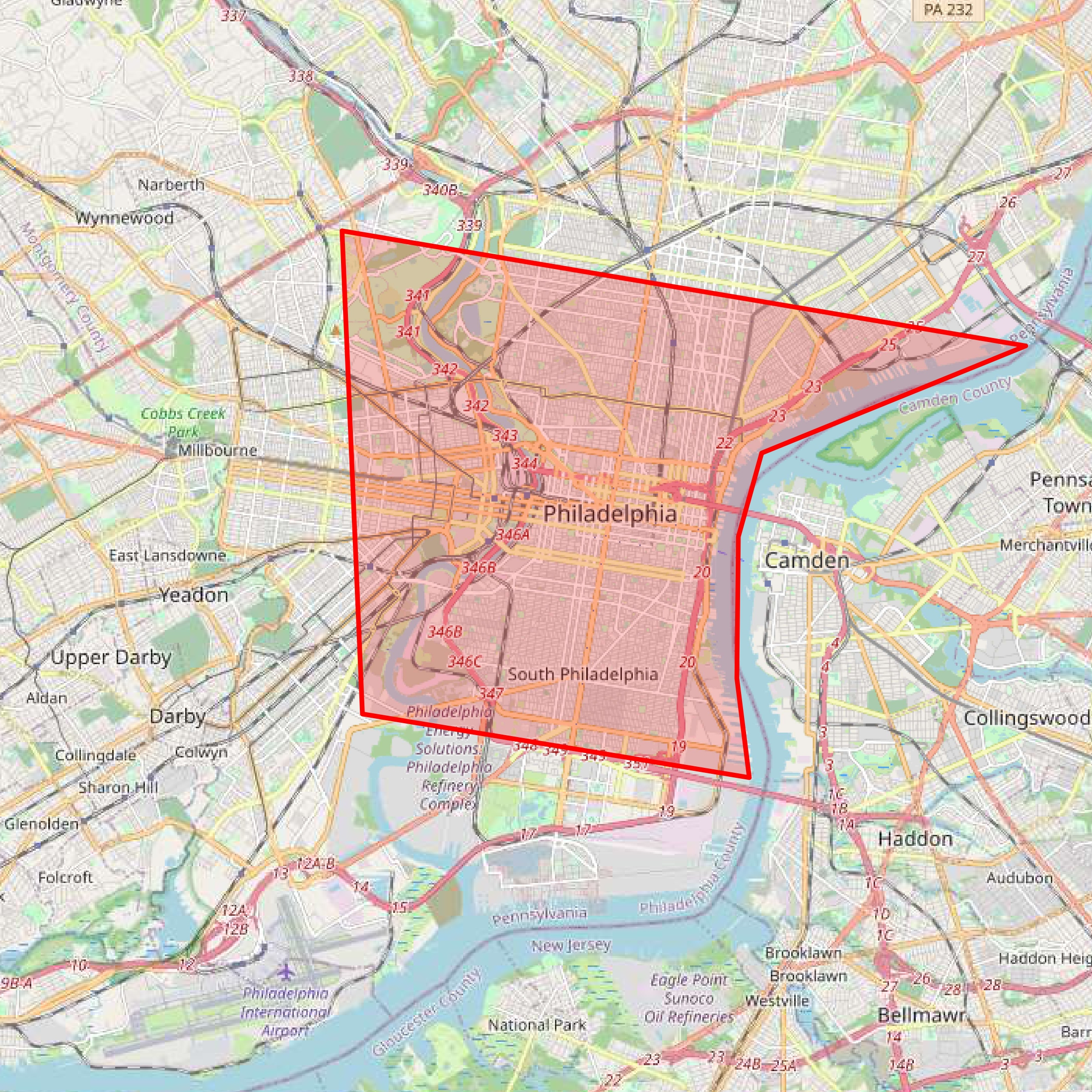}
    \caption{Philadelphia}
    \label{fig:image3}
  \end{subfigure}
  }
  \caption{The RVS instructions are collected over three cities (a–c).}
  \label{fig:cities}
\end{figure*}

\section{Data Collection}
\label{sec:data_collection}

We frame the data collection process as an instructor-follower task, where an instructor needs to communicate to a follower the rendezvous location in relation to the follower’s current location. 
The process is divided into two crowdsourced tasks:  communicating the goal location in writing (here, {\em Instruction Writing}) and following (here, {\em Validation}), corresponding to the two roles -- instructor and follower. Appendix \ref{app:UI} presents a display of the user-interface (UI) of the online assignment.

% \noindent
% We hereby provide the details of the two  UI tasks:

\paragraph{Task 1: Instruction Writing} 
Using the RVS map-graph (Section \ref{sec:task}), we generated the starting points and (within 2km) the respective goal points.  The instructor could view the points on an interactive map with geo-data from OSM, and displayed landmarks along the route, near the goal, in the general area and beyond the route. The goal and nearby landmarks were not shown by their proper names, e.g., instead of `St. Vincent de Paul Church'  the marker displayed `a church'. The instructor could zoom in/out and pan to view the environment. The instructor was requested to describe the location of the goal  in relation to the starting point and landmarks, rather than  providing a step-by-step route description. To prevent easy geolocation by current navigation and geolocation systems, such as street corners, the instructor was restricted to mentioning a maximum of one street by name.

\paragraph{Task 2: Validation} 
% We use a separate task to validate each instruction. 
In this task, the follower is asked to follow the instruction displayed, by pinning the goal location on an interactive map. As the map includes sign symbols of places (e.g., a cross symbol to denote a church), the display also includes
%\reut{would or does include in practic?} 
a legend with the equivalent symbols.
% To ensure that we work with instructions that can be geolocated, an instruction is termed qualified if a follower managed to pin the goal within a 100m distance. The threshold 100m was chosen  as it is the maximum radius of a geoshape from a generated goal in Task 1.
An instruction is considered qualified if the follower pins the goal within 100 meters. This threshold is the maximum radius of the geoshape of the generated goal from Task 1.  Participants were also requested to flag problematic instructions, i.e., those that did not follow the rules in the instruction writing task.  To determine the agreement rate among participants, 50\% of the instructions were validated by at least two participants.

\begin{table*}[th]

\centering
\scalebox{0.86}{
\begin{tabular}{lccccccc}
                    \textbf{City} & \textbf{\begin{tabular}[c]{@{}c@{}}Area \\ Size (km\textasciicircum{}2)\end{tabular}} & \textbf{\begin{tabular}[c]{@{}c@{}}Num. Landmarks\\ in Graph\end{tabular}} & \textbf{\begin{tabular}[c]{@{}c@{}}Num. \\ Instructions\end{tabular}}
                    & \textbf{\begin{tabular}[c]{@{}c@{}}Avg. Path\\ Length (m)\end{tabular}}
                    & \textbf{\begin{tabular}[c]{@{}c@{}}Avg. Text \\ Length\end{tabular}} &
                    \textbf{\begin{tabular}[c]{@{}c@{}}Avg. \\ Entities$^4$\end{tabular}} &
                    \textbf{\begin{tabular}[c]{@{}c@{}}Vocab. \\ Size\end{tabular}}  \\ \hline 
\textbf{Manhattan}  & 32.5                                                                                  & 20,979         
& 8,103  
& 1,098.94                                                                & 43.73  
 & 3.99    
 & 6,365                                                                                                                     \\
\textbf{Pittsburgh} & 34.5                                                                                  & \pz 4,998              & 1,023                                                       & \pz 960.52                                                                  & 41.95  
  & 3.93 
  & 2,195                                                         
\\

\textbf{Philadelphia} & 74.5                                                                                  & 10,302               
 & 1,278   
& 1,096.66                                                                  & 42.96  
   & 3.95  
   & 2,438                                                                                                     
\end{tabular}

}
\caption{ Data Statistics of RVS: statistics over different cities. }
\label{tab:quantitive}

\end{table*}

\begin{table*}[ht]
\centering
\scalebox{0.68}{
\begin{tabular}{lllllllllll}
                                        &           & \multicolumn{2}{c}{\textbf{RVS}} & \multicolumn{2}{c}{\textbf{RUN}} & \multicolumn{2}{c}{\textbf{RxR}} & \multicolumn{2}{c}{\textbf{{\scshape Touchdown} 
}} &                                                   \\
\multicolumn{2}{l}{Phenomenon}                       & \textit{$p$}       & \textit{$\mu$}      & \textit{$p$}               & \textit{$\mu$}                & \textit{$p$}       & \textit{$\mu$}      & \textit{$p$}          & \textit{$\mu$}         & \textbf{Example from RVS}                         \\ \hline
\multicolumn{2}{l}{Proper Names} & 100             & 2              & 100            & 5.96            & 0               & 0              & 0                  & 0                 & \textit{...Duane Reade pharmacy...}                              \\
\multicolumn{2}{l}{Descriptions}            & 96              & 2.48           & 8              & 0.12            & 100             & 8.3            & 100                & 9.2               & \textit{...There is a church across the street...}               \\
\multicolumn{2}{l}{Coreference}                      & 64              & 0.88           & 40             & 0.48            & 64              & 5.3            & 60                 & 1.1               & \textit{...It's on the same block as...}                      \\
\multicolumn{2}{l}{Count}                            & 28              & 0.36           & 8              & 0.08            & 32              & 0.44           & 36                 & 0.4               & \textit{...Southwest of the school are two bicycle parkings.} \\
\multicolumn{2}{l}{Cardinal Direction}               & 96              & 2.2            & 16             & 0.2             & 0               & 0              & 0                  & 0                 & \textit{Go southwest...}                                   \\
\multicolumn{2}{l}{Complex shapes}                & 60              & 1.08           & 44             & 0.76            & 20              & 0.2            & 8                  & 0.8               & \textit{...a block west of the square shaped park...}            \\
\multicolumn{2}{l}{Allocentric Relation}             & 88              & 1.52           & 4              & 0.04            & 76              & 2.4            & 68                 & 1.2               & \textit{...It is west of the bridge...}                          \\
\multicolumn{2}{l}{Egocentric Relation}              & 4              & 0.04            & 76             & 1.36            & 60              & 2.3            & 92                 & 3.6               & \textit{You will pass an Ace Hardware on your left}          \\
\multicolumn{2}{l}{Temporal Condition}               & 8               & 0.08           & 72             & 1.56            & 52              & 0.8            & 84                 & 1.9               & \textit{...Go straight south until you pass the library...}      \\

\multicolumn{2}{l}{Explicit Actions}               & 0               & 0           & 100             & 3.2            & 96              & 0.8            & 100                 & 2.8               & \textit{...Turn left. Continue forward...}      \\
\multicolumn{2}{l}{State Verification}               & 20              & 0.2            & 56             & 0.64            & 84              & 3.1            & 72                 & 1.5               & \textit{...you will see me at the alcohol shop.}               \\
\multicolumn{2}{l}{Negative State Verification}      & 4               & 0.04           & 4              & 0.04            & 0               & 0              & 0                  & 0                 & \textit{...If you see a bike parking, you have gone too far.}  \\ \hline
\multirow{2}{*}{\begin{tabular}[c]{@{}l@{}}Spatial Knowledge \\ \cite{siegel1975development} \end{tabular}}      & Route      & 4               & n/a            & 84             & n/a             & 100             & n/a            & 100                & n/a               & \textit{...turn right on the next street...}                  \\
                                        & Survey     & 96              & n/a            & 16             & n/a             & 0               & n/a            & 0                  & n/a               & \textit{Head east toward the river...}                    
\end{tabular}

}
\caption{Linguistic analysis: we analyze 25 randomly sampled instructions from RVS, RUN, RxR (only instructions given by speakers in the USA), and {\scshape Touchdown}   (only the navigation task).   \textit{$p$} represents the \% of instructions containing the phenomena, while \textit{$\mu$} represents the average number of occurrences within each instruction.}
\label{tab:qualitive}
\end{table*}

\begin{table}[th]

\centering
\scalebox{0.8}{
\begin{tabular}{llll}
\textbf{Feature} &  \textbf{p-value} & \textbf{\begin{tabular}[c]{@{}l@{}}FDR corrected\\ p-value\end{tabular}} & \textbf{F-test} \\ \hline
Num. of entities$^4$                        & 0.56                                                        & 0.56                                                                     & 0.99            \\
Num. of tokens                          & 0.0                                                         & 0.0                                                                      & 2.92            \\
Human distance error                    & 0.0                                                         & 0.0                                                                      & 2.43           
\end{tabular}
}
\caption{One-way analysis of variance (ANOVA) tests were conducted to examine the correlations between goal types and linguistic and human verification features. The p-values were corrected for False Discovery Rate (FDR). A p-value lower than 0.05 indicates a correlation between goal type and a feature.}
\label{tab:anova}
\end{table}

\paragraph{Instructor Training}
 The main challenge of the collection process is training instructors to write high-quality instructions based on survey knowledge (rather than  step-by-step agent-centered descriptions). To address this challenge, the following procedure was implemented: 
 (1) The process starts by  collecting an initial seed of `well-formed' survey-based instructions written by a geospatial expert.
 (2) At least three `well-formed' survey-based knowledge instructions were presented to an unqualified participant one after the other, and the instructor was requested to pinpoint the goal on a map.
 (3) Once the instruction was written by the instructor, it was reviewed by a geospatial expert who provided feedback.
 (4) If a participant successfully produced three well-formed survey-based instructions in a row, the instructor was considered qualified. Every instruction given by a qualified instructor was added to the bank of well-formed survey-based instructions and could be shown to other instructors in training. As more instructors became qualified, the variety of examples increased. 
%  In addition, we randomly sampled instructions throughout the collection process, discarded bad instructions, and provided feedback to the instructors. We also focused on instructors who received low scores in the verification task, where followers pinned the goal location far from the actual goal description, indicating that the instructions were not specific enough to locate the goal. We also focused on instructors that got high scores, as high scores might indicate they gave step-by-step low-level instructions that are easier to follow.
 
\paragraph{Quality Assessment} 
% To ensure the quality of the instructions, we used several methods in addition to the {\em Validation task} and the {\em Instructor Training}. 
We ensured instruction quality by sampling instructions, discarding poor ones, and giving feedback throughout the collection process based on the following criteria: (1) participants who consistently received low distance errors in the verification task (less than 30m average), as it might indicate they gave step-by-step low-level instructions that are easier to follow; (2) instructions that received high distance errors (at least one verification over 2000m); and (3)  instructions from participants who did not participate for over a month.
For participants who failed their reviews (i.e., did not follow the instructions), we reviewed their next three instructions.
% We randomly sampled instructions throughout the collection process, discarded poor instructions, and provided feedback to the instructors.
% We paid close attention to instructors who received low scores in the verification task, where followers pinned the goal location far from the actual goal description, indicating that the instructions were not specific enough to locate the goal. Additionally, we focused on instructors who consistently received high scores, as high scores might indicate they gave step-by-step low-level instructions that are easier to follow.

\section{Data Statistics and Analysis}
\label{sec:stat}

The RVS dataset contains 10,404 validated instructions paired with start and goal coordinates.
The locations are divided among three cities: Manhattan, Pittsburgh, and Philadelphia (Figure \ref{fig:cities} and Table \ref{tab:quantitive}). In the instruction writing task,
146 different participants provided survey-knowledge instructions. 
149 participants completed the validation task, correctly validating 10,404 out of 16,104 tasks (64\%).
% In the validation task, 149 participants completed 16,104 tasks, of which 10,404 (64\%) were correctly validated. 
89\% of validations achieved correct location within 100 meters, indicating high human agreement.
% \reut{89 of what? unclear. agreement criterion also not very clear}
% Analysis of verification accuracy revealed an 89\% human agreement rate based on the successful identification of locations within a 100-meter radius.
% The human agreement rate, i.e., the percentage of verifications that succeeded
% within a 100 meter threshold, is 89\%.    \par

\setcounter{footnote}{4}

\footnotetext{Extracted using  \href{https://chat.openai.com}{ChatGPT} -- https://chat.openai.com}

We conducted a qualitative linguistic analysis of RVS to understand the type of geospatial reasoning required to solve the RVS task. 
We randomly sampled and annotated 25 examples from the Manhattan and Pittsburgh areas of RVS and compared them to previous datasets: {\scshape{RUN}} \citep{paz2019run}, {\scshape Touchdown}  \citep{chen2019touchdown}, and {\scshape RxR} \citep{ku2020room}. % \reut{how did you compare? sampled data from the others and annotated as well? or based their tables?} 
Table \ref{tab:qualitive} details this analysis.
While {\scshape Touchdown}  and {\scshape RxR} contain only mentions of indefinite descriptions, and RUN contains almost exclusively proper names, the RVS dataset contains a relatively balanced use of both descriptions and proper names (not near the goal). This creates a realistic challenge, reflecting the various ways people refer to landmarks.  

Crucially, instructions based on survey knowledge use allocentric rather than egocentric spatial relations.
% require spatial allocentric reasoning characterized by phenomena such as the cardinal directions, complex shapes (e.g., `triangle shaped block'), numerical reasoning, negative state verification, and a tendency to use allocentric rather than egocentric spatial relations.
%However, 
Since {\scshape RxR} and {\scshape Touchdown}  rely on a street/room-level view of the environment  and their participants have only a short time to become familiar with the environment, the instructions contain less spatial allocentric reasoning than RVS.
The RVS dataset displays more allocentric phenomena than the RUN dataset, even though both datasets include a map. 
This is because the RUN dataset encourages participants to use egocentric relations by displaying examples of egocentric relations. Accordingly, as shown in Table \ref{tab:qualitive},  geospatial measures found that RVS contains more survey-based instruction in comparison to the other datasets.

On top of that, {\scshape RUN, RXR}, and {\scshape  Touchdown}  all contain sequential instructions that include many explicit actions and state verifications, making it easier for the model to predict the correct action and verify it after the action is taken.
% by checking for the existence \reut{verified status? verification can exist but be false} of the verification. 
In contrast, the new RVS dataset includes non-sequential instructions with relatively few state verifications and no explicit actions.

%     Since RxR and {\scshape Touchdown}  rely on a street/room-level view of the environment  and their participants have only a short time to become familiar with the environment, the instructions do not exhibit survey knowledge and spatial compositional reasoning, characterized by phenomena such as the use of cardinal directions, complex shapes, negative state verification, and a tendency to use allocentric rather than egocentric spatial relations.  
%  On the other hand, RUN and RVS display a map of the environment, allowing their participants to gain survey knowledge. However, RVS is much richer in  compositional phenomena such as complex shapes (e.g., \enquote*{triangle shaped block}), cardinal directions, allocentric relation, and numerical reasoning (i.e., count). These phenomena which require compositional and configurational reasoning are part of the criteria for differentiating between route and survey knowledge based instruction. Accordingly, as shown in Table \ref{tab:qualitive}, a geospatial expert found that only 16\% of the instructions in RUN reflect survey knowledge, while in RVS 96\% of the instructions contain survey knowledge. 

\begin{table}[t]

\centering
\scalebox{0.93}{
\begin{tabular}{lll}
\textbf{Token} & \textbf{Count} & \textbf{Type}           \\ \hline
Carson         & 65             & street and bridge  \\
Forbes       & 62             & avenue and sport stadium       \\
Pittsburgh           & 54             & city, station and university                   \\
Allegheny         & 29             & avenue                  \\
Smallman         & 23             & street
\end{tabular}

}
\caption{Out-of-Vocabulary Analysis (OOV): Top-5 tokens in the Pittsburgh vocabulary that are absent from the Manhattan vocabulary.}
\label{tab:top-unique}

\end{table}

 \begin{figure*}[th]

  \centering
\scalebox{1}{
            \includegraphics[width=1 \textwidth]{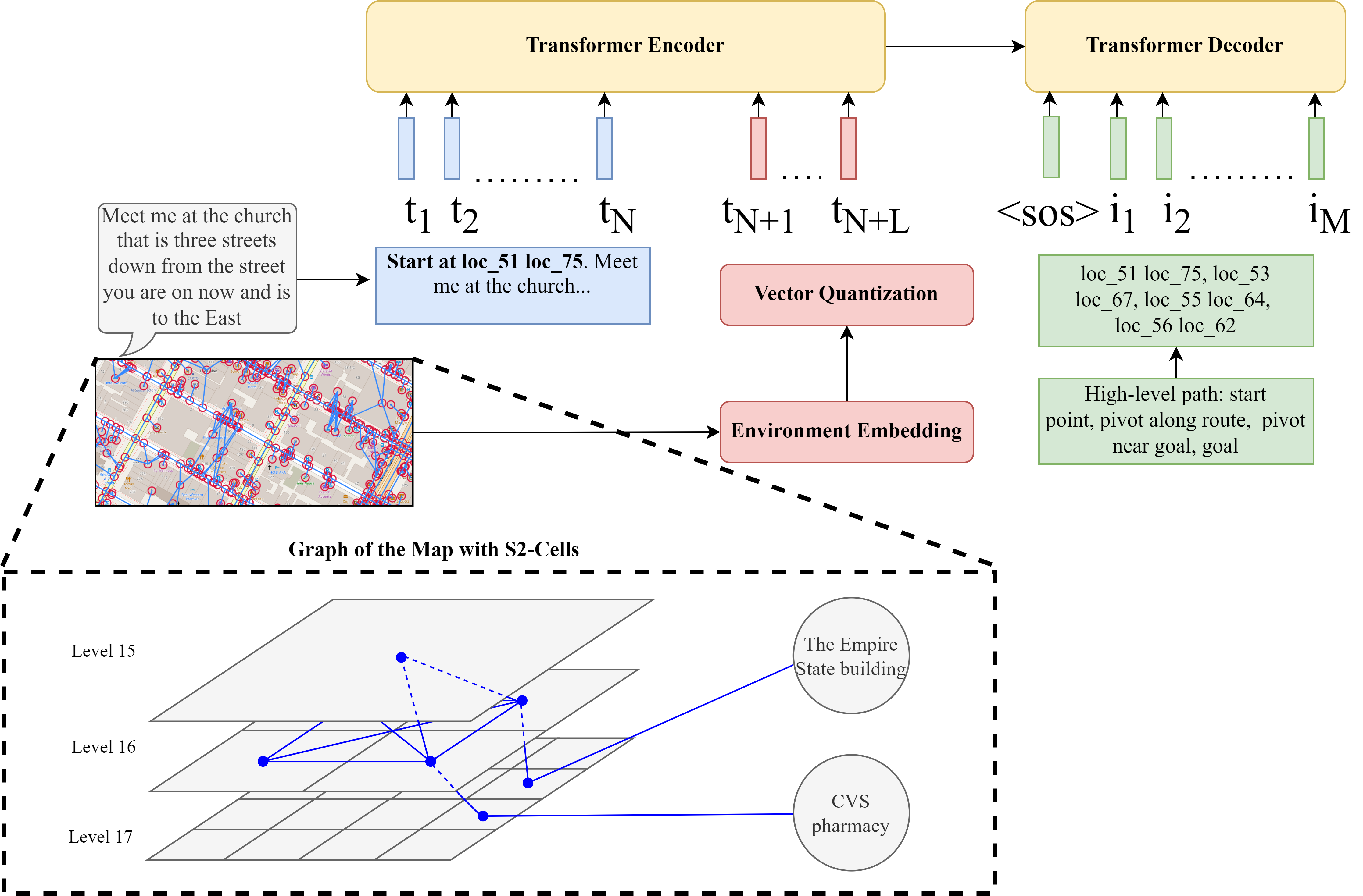}}
            
            \vspace*{-1mm}
                 {\scriptsize{ \begin{spacing}{0.5}

\end{spacing} }}

        \caption
        {The RVS model based on a T5 transformer and a graph representation of the environment. } 

        \label{fig:model}
    \end{figure*}

To prevent simple string-match solutions, the goal location in RVS is always given by its type  (e.g., `restaurant', `parking' etc.) and not by its proper name.  In  Table \ref{tab:anova} we perform 
one-way analysis of variance (ANOVA) tests, to check if there are  entity types easier to locate than others, and if the type affects the instructions. 
%  Table \ref{tab:anova} summarizes the p-values corresponding to the one-way analysis of variance (ANOVA) tests we performed, as well as the False Discovery Rate (FDR) corrected p-values.
%  Two features were found to have a significantly (p<0.05) different distribution between different goal types: the number of entities mentioned in the instructions and the number of tokens in the instructions. The distance error was not found to be significant, indicating that even though the number of entities mentioned and the number of tokens used in the instructions vary depending on the goal type, 
% human ability to geolocate the goal is not affected by its type.
We found that the number of entities and tokens in instructions varied with goal type (p<0.05), but human distance error did not, indicating that human ability to geolocate the goal is not affected by its entity type.

Our out-of-vocabulary (OOV) analysis shows that, unlike  previous  navigation datasets 
%that do not contain proper names of entities 
\citep{chen2019touchdown, ku2020room, anderson2018vision, macmahon2006walk}, RVS presents a challenge with novel entities in a city-split setup, training on one city and testing on a different unseen city. 
Specifically, our analysis  of the vocabularies of two  different cities --- Manhattan and  Pittsburgh --- shows that 36.85\% of the Pittsburgh vocabulary is OOV, i.e., the tokens do not appear in the Manhattan vocabulary. Table \ref{tab:top-unique} shows the top-5 OOV tokens in Pittsburgh. 68\% of OOV tokens are commonly used (82\% of the OOV occurrences) city-specific named entities, like `Carson Street’. Thus, a city-split creates a profound OOV grounding challenge for previously unseen entities.

\section{Models for RVS}
\label{sec:model}

As RVS presents a new multimodal task with unique challenges, we aimed to provide a strong baseline  based on our insights from Section \ref{sec:stat}. We model RVS as a sequence-to-sequence problem, where we map  
the sequence of tokens in the instruction to a sequence of S2-Cells.\footnote{S2Cells are based on S2-geometry, a hierarchical discretization of the Earth’s surface  \citep{hilbert1935stetige}.} %(https://s2geometry.io/)

% \subsection{T5-based models}
% S2-cells are based on S2-geometry \footnote{https://s2geometry.io/} - a robust hierarchical
% discretization of the Earth’s surface based on spherical projections. 
% We describe two models: (1) a T5 Transformer-based model with an encoder-decoder architecture that uses a text-to-text format \citep{raffel2020exploring}; and (2) the {\scshape T5+Graph} model,  which builds upon model (1) by incorporating a graph representation of the environment that is represented on the map. \reut{you need to specify the map signal somewhere -- especially bc of the confusion above} %, to generate the goal location.
% an high-level path from  start point to the goal location.

\paragraph{Encoder}

The encoder encodes the instruction and the starting point's representation. Inspired by \citet{lu2022unified}, who converted pixels to text-based axis locations, we transformed the map’s S2-grid into a two-dimension  discrete  coordinate system (`locX, locY’). The starting point's coordinate is assigned to the S2-Cell containing its geometry. The S2-Cell is linked to an axis position, so the starting position is also assigned an axis position.
% As a result, a location can be encoded as a `locX, locY'. 

\paragraph{Decoder}
Since this is essentially a navigation task without a step-by-step path, we train our model to generate a high-level path, consisting of a sequence of locations starting with the starting point, followed by  prominent landmarks ordered by their directional position from the goal, and ending with the goal. 
We extracted the  prominent landmarks based on the RVS map-graph. As in the encoder, we represent the location in a `locX, locY' format.

\begin{table*}[th]
\centering
\scalebox{0.8}{
\begin{tabular}{lcccccc}
\textbf{Method}   & \textbf{100m Accuracy} & \textbf{250m Accuracy}         & \textbf{Mean Error}                                        & \textbf{Median Error} & \textbf{Max Error}  & \textbf{AUC of Error} \\ \hline
\multicolumn{7}{c}{\cellcolor[HTML]{EFEFEF}\textbf{Manhattan Seen-city Development Results}}                                                                                                                         \\ \hline
\textbf{{\scshape Human} }    & 88.12                   & 95.64                            & \pzz\pz74                                                      & \pzz\pzz4                  & 2,996              & 0.10                  \\
\textbf{{\scshape Stop} }     & 0.00                   & 1.54                            & \pz1,084                                                    & \pz1,124               & 1,929             & 0.41                  \\
\textbf{{\scshape Center} }     & 0.27                   & 1.45                            & \pzz930                                                   & \pzz998              & 1,000            & 0.40                  \\

\textbf{{\scshape Landmark}}     & 0.54                   & 5.26                            & \pzz776                                                   & \pzz815              & 1,384            & 0.39                  \\
\textbf{{\scshape T5} }      

 & \pzz\pzz 27.92 \small{(0.39)}           
% & 0.85 \small{(0.13)}
& \pzz\pzz52.63 \small{(0.45)}
& \pzz\pzz\pz 362 \small{(9)}                                             & \pzz\pzz \pz231 \small{(3)}       & \pzz\pzz2,957 \small{(641)} & \pzz\pzz0.32 \small{(0.00)}           \\

\textbf{{\scshape T5+Graph}} & \pzz\pzz \textbf{29.40} \small{(1.18)}         
% & 0.93 \small{(0.01)} 
& \pzz\pzz \textbf{54.67}  \small{(1.04)}                  

& \pzz\pzz\pz \textbf{357} \small{(7)}                                             & \pzz\pzz \pz \textbf{216} \small{(8)}       & \pzz\pzz3,889 \small{(826)}   & \pzz\pzz0.31 \small{(0.01)}        \\

 \hline

\multicolumn{7}{c}{\cellcolor[HTML]{EFEFEF}\textbf{Pittsburgh Unseen-Development Results}}                                                                                                                      \\ \hline
\textbf{{\scshape Human}}    & 86.94                   & 92.94                            & \pzz\pz99                                                         & \pzz \pz7                     & 2,951               & 0.13                  \\
\textbf{ {\scshape Stop}}     & 0.00                   & 2.05                            & \pzz 960                                                        & \pz 954                   & 1,912               & 0.40                  \\
\textbf{{\scshape Center}}   & 0.00                   & 0.10                            & \pzz 992                                                        & \pz 999                   & \pz999                 & 0.41                  \\
\textbf{{\scshape Landmark}} &    \textbf{1.47}                    &      \textbf{9.48}                           &   \pzz  \textbf{677}                                                       &        \pz  \textbf{691}             &              1,345       &        0.38               \\
\textbf{T5}       & \pzz\pzz 0.49 \small{(1.47)}            &               \pzz\pzz 2.34 \small{(1.44)}                 & \pzz\pzz1,171 \small{(24)}                                                 & \pzz \pz1,107 \small{(14)}            & \pzz\pzz4,701 \small{(101)}         & \pzz\pzz 0.41 \small{(0.00)}           \\
\textbf{{\scshape T5+Graph}} & \pzz\pzz 0.49 \small{(1.01)}            &               \pzz\pzz 2.91 \small{(1.37)}                    & \pzz\pzz1,067 \small{(77)}                                                & \pzz \pz1,039 \small{(56)}            & \pzz\pzz4,102 \small{(727)}         & \pzz\pzz0.40 \small{(0.00)}           \\ \hline

\multicolumn{7}{c}{\cellcolor[HTML]{EFEFEF}\textbf{Philadelphia Unseen-city Zero-shot Results}}                                                                                                                             \\ \hline
\textbf{{\scshape Human}}    & 93.64                   & 97.97                            & \pzz\pz27                                                      & \pzz\pz3                  & 2,708            & 0.05                  \\
\textbf{{\scshape Stop}}     & 0.00                   & 1.80                            & 1,096                                                   & 1,135              & 1,958            & 0.41                  \\

\textbf{{\scshape Center}}     & 0.16                   & 0.47                            & \pz942                                                   & \pz998              & 1,000            & 0.41                  \\
\textbf{{\scshape Landmark}}     & 1.02                   & 7.90                            & \pz707                                                   & \pz713              & 1,384            & 0.38                  \\
\textbf{{\scshape T5}}       & \pzz\pzz0.26 \small{(0.05)}           
% & 0.32 \small{(0.02)} 
& \pzz\pzz1.80 \small{(0.27)}

& \begin{tabular}[c]{@{}c@{}}\pzz\pz1,362 \small{(43)}\end{tabular} & \pzz\pz1,308 \small{(35)}      & \pzz\pzz6,911 \small{(454)}  & \pzz\pzz0.42 \small{(0.00)}           \\

\textbf{{\scshape T5+Graph}} & \pzz\pzz0.31 \small{(0.05)}           
% & 0.39 \small{(0.01)} 
& \pzz\pzz1.93 \small{(0.20)} 

& \pzz\pz1,140 \small{(16)}                                           & \pzz\pz1,161 \small{(8)}       & \pzz\pzz5,277 \small{(372)} & \pzz\pzz0.41 \small{(0.00)}          
\end{tabular}
}
\caption{Results over the test and development sets. The distance errors are presented in meters. For the learning models, we report the mean over three random initializations and the standard-deviation (STD) is in brackets. }
\label{tab:results}

\end{table*}

\begin{table*}[th]
\centering
\scalebox{0.81}{

\begin{tabular}{lccccl}
                     \textbf{Split}      & \textit{$p$} & \textbf{Min $c$} & \textbf{Max $c$} & \textbf{Avg. $c$} & \textbf{Example from RVS}                                                                                                                                                                                       \\ \hline
Seen-City   & 61                                              & 3                                           & 9                                           & 5.4                                          & \multirow{2}{*}{\begin{tabular}[c]{@{}l@{}}\textit{I am \underline{{ northeast of you}} \underline{\underline{at a toilet}} \underline{{near the corner of Bayard Street}}.  \underline{{ To }}}  \\ \textit{\underline{{its south is a park}} and \underline{{the Louis J. Lefkowitz State Office Building}}...}\end{tabular}} \\
Unseen-City & 13                                              & 2                                           & 8                                           & 5.05                                         &                                                                                                                                                                                                               
\end{tabular}

}
\caption{Spatial relations analysis of 20 samples. $c$ and \textit{$p$} represent the number and percentage of spatial relations to the location predicted  by {\scshape T5+Graph} that match those mentioned in the text, respectively. 
In the examples, the matched relations are underlined, and the unmatched relations are double-underlined.
%$F/C$ is the percentage of conditions fulfilled in an instruction.
}
\label{tab:error_seen2}

\end{table*}

\begin{table}[th]
\centering
\scalebox{0.91}{

\begin{tabular}{lcc}
\textbf{Type of Pred. and True Goal Relation}                         & \multicolumn{1}{l}{\textbf{$p$}}
  & \multicolumn{1}{l}{\textbf{$c$}}
\\ \hline
On the same S2-Cell            & 25  &    5                    \\
Same cardinal-direction from start point & 95  & 19                            \\
On the same street             & 45    & 9                         \\
Have the same type of entity     & 50     & 10                       
\end{tabular}

}
\caption{Error analysis of 20 instructions and their corresponding {\scshape T5+Graph}  results in the seen-city split. $c$ and \textit{$p$} represent the number and percentage of the instructions that contain the types of relation between the predicted  goal  and the true goal.  }
\label{tab:error_seen1}

\end{table}

\paragraph{The World as a Graph}
\label{sec:graph}

 A location can be represented by its position ({\em where} the location is) or by its semantics ({\em what} is present at the location, e.g., `a bar'). Semantic knowledge is crucial for grounding mentioned entities to their physical references in the environment. 
To this end, we aim to connect the semantic and positional knowledge using a novel RVS map-graph. The RVS map-graph is a heterogeneous graph containing location nodes (semantic) and S2-cell nodes (positional). First, we connected each location node to its smallest containing S2-cell (see Figure 3), also instantiating each S2-cell as an independent node in the graph. Then, as the S2-geometry is a hierarchical structure, we add both within-level and between-level edges between S2-cell nodes. Specifically, we connect each S2-cell to its immediate neighbors at the same level, and we connect each S2-cell to its containing S2-cell at the next level up in the hierarchy (see Figure \ref{fig:model}). To learn a joint embedding space for locations and S2-cells, we compute random walks on the graph using node2vec algorithm \citep{grover2016node2vec}.
% \paragraph{Learned Vector Quantization}
% Following \citet{yu2021vector}, we use a linear projection to cluster the graph embeddings into K categories  using the k-means algorithm with cosine similarity distance.  A new token is assigned to each  category and added to the tokenizer's vocabulary. The quantization token is passed with the instruction's tokens to the transformer encoder.
Following \citet{yu2021vector}, we use linear projection to cluster the graph embeddings into K categories  using the k-means algorithm with cosine similarity distance.  A new token is assigned to each  category and added to the tokenizer's vocabulary. We perform multiple clusters and pass the graph's tokens with the instruction's tokens to the transformer encoder.

%  \begin{figure*}[th]

%   \centering
% \scalebox{1}{
%             \includegraphics[width=1 \textwidth]{img/graph.png}}
            
%             \vspace*{-1mm}
%                  {\scriptsize{ \begin{spacing}{0.5}

% \end{spacing} }}

%         \caption
%         {The environment graph structure representing   } 

%         \label{fig:graph}
%     \end{figure*}

\section{ Experimental Setup}

\paragraph{Evaluation} 

We use six evaluation metrics: (1) 100m accuracy, the task is considered completed if the agent is within a 100m distance from the goal; (2) 250m accuracy for coarse-grained accuracy evaluation; (3) mean distance error; (4) median distance error; (5) maximum distance error; and (6) area under the curve (AUC) distance error.

\paragraph{Setup and Data-Split} 
We use a zero-shot (ZS) city-based split, where we train on one city, validate on a second city, and test on a third city.
Specifically, RVS's setup consists of
(i) a \textbf{training-set} containing 7,000 instructions from Manhattan; (ii) a \textbf{seen-city development-set} containing 1,103 instructions from Manhattan;  (iii) an \textbf{unseen-city development-set} containing 1,023 instructions from Pittsburgh; and (iv) a \textbf{test-set}  containing 1,278 instructions from Philadelphia. The ZS split raises profound challenges (e.g., OOV) at inference time, as described in Section \ref{sec:stat}.   

\paragraph{Learning} We use supervised learning by maximizing the log-likelihood of high-level paths. We train the model with AdamW \citep{adamW} for optimization. Details of the learning and hyperparameters
are provided in Appendix \ref{app:exp}.

\paragraph{Systems} We evaluate three non-learning baselines: (1)~{\scshape Stop}: predicts the starting point as the goal location; (2)~{\scshape Center}:  predicts the closest location towards the center of the region within a 1000-meter radius from the starting point; (3)~{\scshape Landmark}: predicts the location of a prominent landmark in the map within a radius of 1000 meters. A landmark is considered prominent if it has one of the following tags (appearing in descending order of importance): (a) Wikipedia page; (b) Wikidata page; (c) a part of a brand; (d) a tourist attraction; (e) an amenity; and (f) a shop. 

We also evaluate two learning models described in Section \ref{sec:model}. The first model is based on {\scshape T5}, and  the second model  {\scshape T5+Graph}, is based on  {\scshape T5}   with an addition of a graph-based representation of the environment. This representation  is described in Section \ref{sec:graph} and depicted in Figure \ref{fig:model}.

\section{Results}

Table \ref{tab:results} shows the seen-city development, and unseen-city zero-shot (ZS) results for our six evaluation metrics. The human performance provides an upper bound for the RVS task performance, while the simple {\scshape Stop} is a simple lower bound baseline. 
Although the {\scshape T5+Graph}  outperforms the non-learning baselines ({\scshape Stop}, {\scshape Center}, and {\scshape Landmark}) in the seen-city split, there is still a gap of  58.72\% and 40.97\% in the 100m and 250 accuracies, respectively. 
The {\scshape Landmark} model outperforms other non-learning models, suggesting that the goal location is more likely to be around prominent landmarks than in other areas. 

Despite the 2km maximum distance between the start and goal, we did not  constrain our models or teach them S2-Cell distances. 
%Real-world paths could be longer. 
So the maximum error of the learned models was greater than 2km.  
% We evaluate our model on RVS and assess the  contribution of the added graph component.
The improved performance of the {\scshape T5+Graph} over the {\scshape T5} indicates  
% The graph component improves the performance of the {\scshape T5} and model, proving 
that the added graph can capture semantic geospatial information. 

The novel ZS city-split setup we introduced 
%in previous navigation tasks. The setup is a hard 
provides a profound challenge for natural language understanding due to the appearance of new spatial relations and new entities in the environment. This can be seen in the inability of the  learning model to generalize from seen to unseen environments, resulting in low performance, even underperforming the  non-learning {\scshape Landmark} baseline.
% , with a gap of 1\% in the 100m accuracy.
%and 6\% in the 100m and 250m accuracies, respectively. 

%Handling new environments is a realistic challenge which raises the bar on models performance, requiring generalization into new environments.   

% This is mainly due to change of the environment and the entrance of many OOV into the descriptions. 

% Despite this progress, 
% %made by the {\scshape T5+Graph} model, 
% there is still a considerable gap between the {\scshape T5+Graph} performance and that of {\scshape Human}, particularly in the ZS split. One potential solution to this issue, which warrants further research, could be to generate synthetic data for fine-tuning the model on a new city. \par

% Following Table \ref{tab:anova}, we conducted an ANOVA test on the correlation between the goal type and the distance error of the {\scshape T5+Graph} model. The results -- p-value of 0.34, indicates there is no correlation between the type of the goal and the performance of {\scshape T5+Graph}. \par

%\paragraph{Error Analysis}
Tables \ref{tab:error_seen2} and \ref{tab:error_seen1} show an  error analysis of 20 examples of the {\scshape T5+Graph}'s results in seen-city and unseen-city splits. 
As shown in Table \ref{tab:error_seen2}, the model must consider multiple spatial relations to handle RVS.
\footnote{A comparative analysis of 20 RXR instructions revealed that up to two spatial relations per navigational step necessitate reasoning for successful completion.}
However, the model only successfully manages to predict a goal that matches the spatial relations mentioned in the text in 61\% and 13\% for the seen-city and unseen-city splits, respectively. Table \ref{tab:error_seen1} shows that in the seen-city split, the model correctly identifies the cardinal directions in most cases, suggesting that it has learned the outline configuration of the map. 
In half of the cases, the model correctly identifies the type of the entity. The model correctly identified the street in 45\% of cases, and in 88.89\% of those cases, the street was mentioned by name in the text. This is lower than the 90\% of all sampled instructions that mentioned street names, suggesting that simply mentioning a street by name is not sufficient for the model to correctly produce a location on that street. In 25\% of the cases, the granularity of the S2-Cells is not high enough to distinguish between the predicted and true goal, suggesting that a higher level of S2-Cell could reduce these cases.

Following Table \ref{tab:anova}, we conducted an ANOVA test and found no correlation between goal type and distance error for T5+Graph (p-value = 0.34).
% Table \ref{tab:error_unseen} provides a fine-grained analysis of the error categories for the ZS split of 20 samples. In most cases, the {\scshape T5+Graph} model prefers to predict a location close to the start point -- almost the exact location ({\scshape Stay}  ), the closest prominent landmark ({\scshape Prominent-Landmark}) or a close location that fulfills one spatial relation mentioned in the text ({\scshape 1-Condition}). Unlike the non-learning ({\scshape Landmark}) model, in  {\scshape Prominent-Landmark} the model predicts a location near the closest landmark to the start point and not the most prominent landmark (e.g., has a Wikipedia page) within a 1000 meters. The {\scshape Prominent-Landmark} suggests that the model learns that the goal is more likely to be where there are `interesting' landmarks. The {\scshape 1-Condition} suggests that the model lacks the ability to reason over multiple spatial conditions which is required in RVS. 

\section{Related Work}

As people move 
% through their environment,
they perceive their surroundings and acquire knowledge
of the space, known as cognitive mapping \citep{tolman1948cognitive}. One influential cognitive mapping theory \citep{siegel1975development} divides cognitive mapping ability into three levels.
Landmark knowledge, consisting of landmarks (e.g., mountains and buildings) and their attributes (e.g., location, size, color), 
%and is often used to anchor geospatial instructions. 
%Route knowledge
Route knowledge,  altered by the  traveler's  changing viewpoint  \citep{taylor1992descriptions, taylor1992spatial,taylor1996perspective} %.the routes between locations 
%and encodes a sequence of procedures for way-finding. This sequence can be 
and coded directly (e.g., \enquote{turn right, then straight}) \citep{tlauka1994effect}, or as condition-action rules based on landmark-direction associations
%, where nodes correspond to the main landmarks acquired in the landmarks knowledge stage, 
(e.g.,  \enquote{turn right at the church, then straight} \citep{kuipers1978modeling, thorndyke1981spatial}),  and survey knowledge,  where people form a `cognitive map' of the environment, an overview of the geospatial layout, and gain awareness of relationships between different geospatial components, even outside the route. Survey knowledge is independent of a person's own position, and enables her to form  different routes, refer to cardinal directions, describe landmarks at different resolution levels, 
%, e.g., \enquote{the church is at Chelsea neighborhood, on the block intersection of 23rd and 7th.}
and describe complex shapes of abstract features such as `blocks'. Such information is less likely to be acquired from direct experience in the environment, but is portrayed on maps \citep{thorndyke1982differences}.
Thus, instructions based on such knowledge mirror the complex understanding of the environment.

% The process of grounding in NL requires a representation of the world that the utterance refers to, and this representation

In grounded NLP tasks, participants acquire knowledge over an environment provided with the task. This environment can be based on different sources, most commonly visual sensors  with real \citep{Qi_2020_CVPR,blukis2018following,wang2018look} or synthetic imagery \citep{yan2018chalet, misra2018mapping,shridhar2020alfred}. % People use different procedures to make geospatial judgments, depending on the type of knowledge they have. 
In a visual environment, participants travel through the environment, view it from a point on the ground that is on the same plane as the objects, and acquire route knowledge. 
%of the routes connecting the diverse locations.
\citet{thorndyke1982differences} found that subjects who learned an environment by walking through it were limited to route-based knowledge and used egocentric spatial relation expressions (e.g., `on your right') in their instructions. This observation was reinforced by \citet{chen2019touchdown} analysis of {\scshape Touchdown}  \citep{chen2019touchdown} and {\scshape R2R} \cite{anderson2018vision} — two navigation tasks with  walk-through environments.
% \reut{is RxR and R2R same thing?}
% Participants describe the world from their current view point use directional terms such as \enquote*{on your right} and \enquote*{forward}, while referencing to landmarks relative to their current position in the world.
% Moreover, with only moderate exposure to the environment walk-through  tasks are inferior to map-based tasks, such that the participants are at an initial state of cognitive mapping and their knowledge of the current environment is primitive, and therefore they cannot give instructions beyond a route-based knowledge \citep{thorndyke1982differences}. 
% let alone an instruction based on a bird's eye view of the environment . 
% Thus, instructions like \enquote{3 blocks from...} are very infrequent. \par

% Previous datasets for NL navigation based on a symbolic world representation,  HCRC \cite{anderson1991hcrc,vogel2010learning, levit2007interpretation} 
Another type of environment uses maps \citep{anderson1991hcrc, paz2019run, vogel2010learning, levit2007interpretation, vasudevan2021talk2nav, de2018talk}, where instructors can view the environment from above and gain survey knowledge of global geospatial relations. However, previous works with maps have either presented small, simplistic environments \citep{anderson1991hcrc, de2018talk} 
or the task's setup has encouraged participants to give egocentric sequential instructions limited to the route 
% e knowledge as they are given a specific route and shown examples of route knowledge based instructions
\citep{paz2019run, de2018talk, vasudevan2021talk2nav}.
% and/or the followers moves in a visual environment that is new to them, thus, landmarks' references are limited to the route only \citep{ de2018talk, vasudevan2021talk2nav}.
In contrast, RVS focuses on instructions that encode survey knowledge and require configurational and allocentric reasoning over a large, entity-dense environment. 

There are sharp differences  between indoor \citep{ku2020room, anderson2018vision} and outdoor \citep{chen2019touchdown, paz2019run, de2018talk, vasudevan2021talk2nav, anderson1991hcrc} navigation instructions. 
Indoor environments contain many entities referred to as definite descriptions (e.g., `the chair') and few landmarks that can be referred to by their proper name
%\footnote{Known as {\em named entities} in NLP and as {\em rigid designators} in formal semantics \cite{kripke1972naming}.} 
(`The Blue Room in the White House'). In outdoor environments, people tend to mix the use of proper names (e.g., `the Empire State building') and definite descriptions (e.g., `the school'). However, previous outdoor navigation tasks either  contain only  definite descriptions  \citep{chen2019touchdown, vasudevan2021talk2nav} or almost exclusively proper names \citep{paz2019run}.
RVS contains a balanced amount of both.

\section{Where Do We Go From Here?}

\paragraph{Bridging the Human-AI Performance Gap} A substantial gulf separates current models' performance from human performance in the RVS task. In seen environments, models lag behind by 58.72\% in 100-meter accuracy and 212 meters in median error. This gap widens further in unseen environments, with a staggering 93.33\% difference in 100-meter accuracy and 1,158 meters in median error.  The challenge of bridging this gap could unlock thrilling research avenues  that push the boundaries of this  task.

\paragraph{Spatial Large Language Models}
One promising approach to tackle this challenge lies in the development of spatial large language models (LLMs) specifically pre-trained for geolocation based on textual descriptions. 
% \reut{how does such a thing look like? can you give an intuition? generating embeddings for location coordinates? for the language? for a mix of both?}
Such models could unlock the vast potential of textual geospatial information readily available online \citep{spink2002sex, sanderson2004analyzing}. They could empower natural language-driven geospatial queries and support Geo-Information Retrieval (GIR) processes. Additionally, generating instructions that describe a location based on relative landmarks – rather than explicit actions like `turn right', which are not always relevant or sufficient for navigation in many parts of the world --- can enable people to follow instructions which are less `robotic’, more natural, and more relevant. Looking beyond navigation, spatial LLMs could also play a crucial role in enhancing the accessibility and usability of geospatial data. By enabling users to interact with maps and spatial information using natural language, LLMs can bridge the gap between human language and spatial data representations, making these resources more accessible to a wider range of users.

\paragraph{Seeing the Streets: Integrating Visual Cues} Humans perceive the world through different signals (e.g.,  images and  sounds) that they get from their senses. Similarly, to understand the world, artificial intelligence research also tries to solve problems that use multimodal data \citep{antol2015vqa, paz2020zest, ji2022abstract}. While maps are one modality that can be used in navigation, it is interesting to note that regions of the maps can be augmented by street view images, such as Google Street View imagery,\footnote{StreetLearn dataset \cite{mirowski2019streetlearn} contains images for the Manhattan and Pittsburgh regions in RVS.} to integrate the visual modality in the RVS dataset. Alternatively, the RVS dataset represents maps as symbolic world representations, which do not account for the visual perception of maps by humans. Therefore, it would be interesting to use image representation instead of graphs in the RVS dataset.\footnote{The GitHub repository for the RVS dataset contains maps’ imagery, which can be accessed at the following link:  \url{https://github.com/OnlpLab/RVS}} Visual descriptions that appear in RVS, like the shape of a "triangular block" are far more evident in images than in the symbolic map representation.

% \paragraph{Closing the Performance-Gap } There is a huge performance gap between the models and the human performance absolute difference--  58.72\% in 100 meter accuracy and 212 meters in the median error in the seen-environment, and an even larger gap on the unseen-environment -- 93.33\%  in 100 meter accuracy and 1,158 meters in the median error. Closing the performance gap with humans unlocks exciting frontiers for future research in this task.

\section{Conclusion}
This work presents the RVS task and dataset, which present a new focus on understanding geospatial instructions based on survey knowledge of urban environments. Our analysis shows that the data presents 
profound spatial-reasoning challenges such as allocentric relations,  multiple relations, cardinal directions, and more,   requiring models with
novel representations of the environment that can enhance and complement the language understanding capacity of LLMs. Our  results show that our  zero-shot city split set-up presents a major  challenge, leaving ample space for further research on this benchmark and task. 
% \reut{edited, check}

\section*{Limitations} 

% \reut{I dont understand why this write up is a limitation, and limitation of what. Can you preface or end it with a take-home statement of the limitation and what could have been done otherwise if needed?}\reut{Also, I find it hard to believe that this is the only limitation. Different limitations are only taking these 3 cities and not other smaller urban or rural areas. Another limitation can be lack of height dimension? another can be the language aspect (and maybe ref HEGEL as a follow up addressing it? We need more, and more convincing, content here. I think it does not count against page limit - does it?}

In the data collection process (described in Section \ref{sec:data_collection}) we showed participants an interactive map with the start and goal points, as well as landmarks along the route, near the goal, and in the general area beyond the route.  One of our guidelines for collecting the data is to allow participants to use a mix of proper names and definite descriptions without giving the location of the goal by mentioning proper names adjacent to it, so that a named entity recognition (NER) system would not   be able to locate the goal.
To enforce this guideline, we displayed the landmarks with different levels of information: for landmarks near the goal (less than 200m), we displayed partial information, excluding the proper name; for landmarks far from the goal (more than 200m), we displayed all the information.
For example, for a landmark of a restaurant with the tag name `Kofoo', we displayed multiple tags without the tag name if it was located near the goal: `amenity: restaurant, cuisine: `korean'. This allowed the participant to refer to `Kofoo' as a `restaurant` or a `korean restaurant'.
To achieve this, we displayed pop-up markers of the landmarks and requested the participants to provide the instructions  using only descriptions of landmarks in the pop-up markers (see Appendix \ref{app:UI}). 
While aiming to minimize information overload (IO), our study presented only 40 of these landmark pop-up markers on the map. 
Landmark selection prioritized prominence based on pre-defined tags like "wikipedia" and "brand." However, this approach restricted user choice and potentially introduced bias. In dense areas like Manhattan, showcasing merely 40 landmarks concealed 99.81\% of potential reference landmarks. Moreover, relying solely on specific tags may have neglected other prominent features readily used for navigation, such as easily identifiable landmarks on street corners. This potential mismatch between presented and naturally chosen landmarks could have influenced navigational accuracy. While increasing the displayed landmarks seems intuitive, it could exacerbate IO and prolong search times for relevant landmarks. Thus, the challenge lies in striking a balance between minimizing IO and providing sufficient landmarks for accurate wayfinding.

\section*{Acknowledgements}

This research has been funded by the European Research Council (ERC), grant number 677352 and by a grant from the Israeli Science Foundation (ISF) number 670/23, for which we are grateful. The research was further supported by
a KAMIN grant from the Israeli Innovation Author-
ity, and computing resources kindly
funded by a VATAT grant and via the Data Science
Institute from Bar-Ilan University (BIU-DSI). 
We are also grateful for the additional support provided by a Google grant.

\bibliography{main}
\bibliographystyle{acl_natbib}

\appendix

\section{Data Collection Details}

\paragraph{Participants}
We collected the RVS dataset using
\href{https://www.mturk.com/}{Amazon Mechanical Turk (MTurk)}. 
We did not collect any information that could be used to identify the participants. We presented the task to the participants as part of a research on navigation instructions. 
We worked with both past MTurk workers and new workers who had a  99\%  percentage assignment approval rate and at least 500 approved HITs. Only English speakers were allowed to participate. The base pay was \$$0.40$ for writing instructions and \$$0.15$ for completing a validation task. Instead of giving bonuses based on successful validation, we rewarded workers who generated high-quality instructions based on survey-knowledge that met our criteria, such as not mentioning more than one street by name. After evaluating worker performance through random sampling of instructions, we offered bonuses ranging from  \$$0.5$ to \$$2.0$  to those who performed well. 
All but three of the 149 participants who took part in the validation task also participated in the instruction writing task.

\paragraph{Instructions vs. Descriptions}
Although our `instructions' are non-sequential and thus  differ from typical instructions in previous navigation tasks \cite{paz2019run, chen2019touchdown, ku2020room}, we chose the term `instruction' and not `description' for the following reasons: (1) The term `descriptions' is used in geolocation tasks where place descriptions are given \citep{hegel}. Unlike RVS, in geolocation tasks there is no assumption for a starting point \citep{krause2020deriving,krause2023geographic}. In RVS, we give instructions on how to find point B, given point A as a starting point. (2) Instructions are usually sequential, but they don’t have to be (e.g., a set of assembly instructions for a toy is non-sequential because the steps can be followed in any order and still result in a completed toy).

\begin{figure}[t]
\centering
\scalebox{0.48}{
\includegraphics[width=\textwidth]{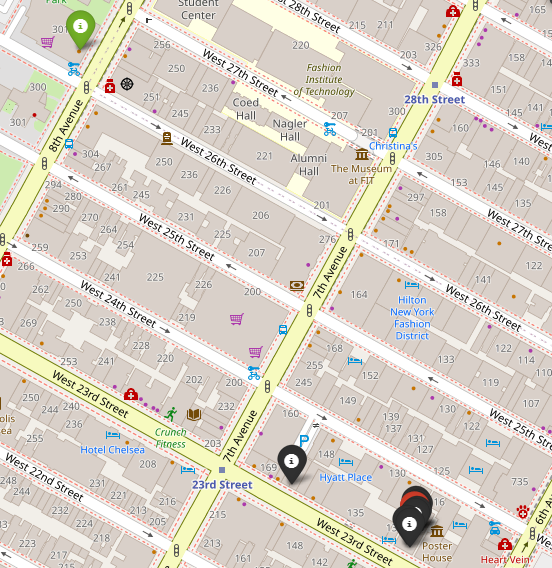}}
  {\footnotesize{

                 \begin{spacing}{0.5}
                
                 \begin{flushleft} 
 \begin{mdframed}
 
\textit{
\textbf{Meet me at the church on West 23rd street. It will be southeast of your current location. On your way, you will see Shirley Goodman Resource Center, which is three blocks north and half a block west of the church. The church is in the middle of the block, if you reach a college you have gone too far.
  } 
 }
 \end{mdframed}
 \end{flushleft} 
\end{spacing} }}
 \caption{
 Example of Multiple Validations: (i) starting point (green marker), (ii) goal (red marker), and (iii) predicted goal by participants (black markers).
 }
\label{fig:verify_multiple}
\end{figure}

\paragraph{Multiple Validations} 
 In order to determine the agreement rate among participants, at least two participants validated 50\% of the instructions, as shown in Figure \ref{fig:verify_multiple}.

\paragraph{Selection of Cities}
The study selected three cities to create a realistic scenario where training is done on one city and testing is done on another. Manhattan was selected as the training set because it is the most entity-dense environment and will allow for maximum unique paths. Additionally, Manhattan and Pittsburgh were chosen because the StreetLearn dataset \cite{mirowski2019streetlearn} released Google Street View imagery for these areas, which might allow future integration of images.

\begin{table*}[h]
\centering
\scalebox{0.8}{

\begin{tabular}{lcccccc}
\textbf{Model}  & \textbf{100m Accuracy} & \textbf{250m Accuracy}         & \textbf{Mean Error}                                        & \textbf{Median Error} & \textbf{Max Error}  & \textbf{AUC of Error} \\ 
 \hline

\multicolumn{7}{c}{\cellcolor[HTML]{EFEFEF}\textbf{Train on  Pittsburgh}}                                                                                                                      \\ \hline

\textbf{{\scshape T5}} & \pzz\pzz0.00            
% & 0.39 \small{(0.01)} 
& \pzz\pzz  1.09

& \pzz\pz1,085                                        & \pzz\pz1,119        & \pzz\pzz1,969  & \pzz\pzz0.41   

\\

\textbf{{\scshape T5+Graph}} & \pzz\pzz0.18            
% & 0.39 \small{(0.01)} 
& \pzz\pzz2.45 

& \pzz\pz1,219                                        & \pzz\pz1,172        & \pzz\pzz5,954  & \pzz\pzz0.41   

\\

\multicolumn{7}{c}{\cellcolor[HTML]{EFEFEF}\textbf{Train on  Philadelphia}}                                                                                                                      \\ \hline

\textbf{{\scshape T5}} & \pzz\pzz0.00            
% & 0.39 \small{(0.01)} 
& \pzz\pzz  1.54

& \pzz\pz1,085                                        & \pzz\pz1,124        & \pzz\pzz1,929  & \pzz\pzz0.41   

\\

\textbf{{\scshape T5+Graph}} & \pzz\pzz 0.27            
% & 0.39 \small{(0.01)} 
& \pzz\pzz1.72 

& \pzz\pz1,869                                        & \pzz\pz1,232        & \pzz\pzz7,436  & \pzz\pzz0.42   

\\
   
\end{tabular}

}

\caption{Results for testing on Manhattan using different training sets from Pittsburgh or Philadelphia. }
\label{tab:results_permutation}

\end{table*}

\paragraph{Path Length Limitation} To ensure accurate, precise, and geolocatable navigation instructions for participants, we implemented a two-kilometer radius limitation. Our preliminary experiments with MTurk participants revealed that they experienced difficulties in finding the goal location when the distance between the start point and the goal exceeded two kilometers. Additionally, the RVS dataset is designed to facilitate high-granularity urban geolocation, making it essential to restrict the navigation range to a manageable distance.

\section{{\scshape T5}-based models}

\paragraph{The Graph Embedding} The graph was constructed using three levels of S2-Cells: 15, 16, and 17. At level 16, each sub-graph consisting of four neighboring S2-Cells was fully connected. All S2-Cells in the graph were linked to their parent S2-Cell based on the S2-geometry's hierarchy (i.e., level 17 S2-Cells were connected to level 16 S2-Cells and level 16 S2-Cells were connected to level 15 S2-Cells). Extracted entities from OSM and  \href{https://www.wikidata.org}{Wikidata}\footnotetext{Wikidata is a free and open knowledge base that acts as central storage for structured data of its Wikimedia sister projects, including Wikipedia, Wikivoyage, Wiktionary, Wikisource, and others} were linked to the smallest level 17 S2-Cell that encompassed their geometry. The node of the entity included additional data such as its geometry, type, and name. Random walks on the graph were performed using node2vec \citep{grover2016node2vec}.

\paragraph{Experimental Setup Details}

For both T5-base models, we use a pre-trained `T5-Base' model from \href{https://huggingface.co/transformers/v3.0.2/_modules/transformers/modeling_tf_t5.html#TFT5ForConditionalGeneration}{Hugging Face Hub}, which is licensed under the Apache License 2.0. The T5 model was trained on the Colossal Clean Crawled Corpus (C4, \citet{raffel2020exploring}).
The cross-entropy loss function was optimized with
AdamW optimizer \citep{adamW}. The hyperparameter tuning is based on the average results
run with three different seeds. We used a learning rate
of 1e-4. The S2-cell level was searched in [15,
16, 17, 18] and 16 was chosen. The number of clusters for the quantization process was searched in [50, 100, 150, 200, 250] and 150 was chosen. We used 2 quantization layers. Number of epochs for
early stopping was based on their average learning
curve. 
We used the following parameters for the node2vec algorithm: an embedding size of 1024, a walk length of 20, 200 walks, a context window size of 10, a word batch of 4, and 5 epochs.

\subsection{S2-Geometry}
S2Cells are a hierarchical discretization of the Earth's surface, enabling efficient representation and computation of geospatial data. S2Cells are based on S2-geometry a mathematical framework for representing and computing shapes on the sphere \cite{hilbert1935stetige}. 
Each cell is a quadrilateral bounded by four geodesics (shortest path between two points on a curved surface). 
The top level of the hierarchy is obtained by projecting the six faces of a cube onto the unit sphere, and lower levels are obtained by subdividing each cell into four children recursively.  S2Cells are globally uniform, i.e., all of the cells at the same level have the same size and shape, regardless of where they are located on the Earth's surface. The level is defined as the number of times the cell has been subdivided (starting with a face cell). Cells levels range from 0 to 30. The smallest cells at level 30 are called leaf cells; there are $6 * 4^{30}$ of them in total, each about 1cm across on the Earth’s surface.

\label{app:exp}

% The cross-entropy loss function was optimized with Adam optimizer \citep{Adam}.
% The hyperparameter tuning is based on the average results run with three different seeds.
% The Learning rate was searched in [1e-5, 1e-4,  1e-3] and a 1e-5 was chosen.
% The S2cell level was searched in [15, 16, 17] and 16 was chosen. 16 level has an edge length between 140  and 153 meters, which is 
% Number-of-epochs for early stopping was based on their average learning curve.

\begin{figure*}[h]
\centering
\scalebox{1}{
\includegraphics[width=\textwidth]{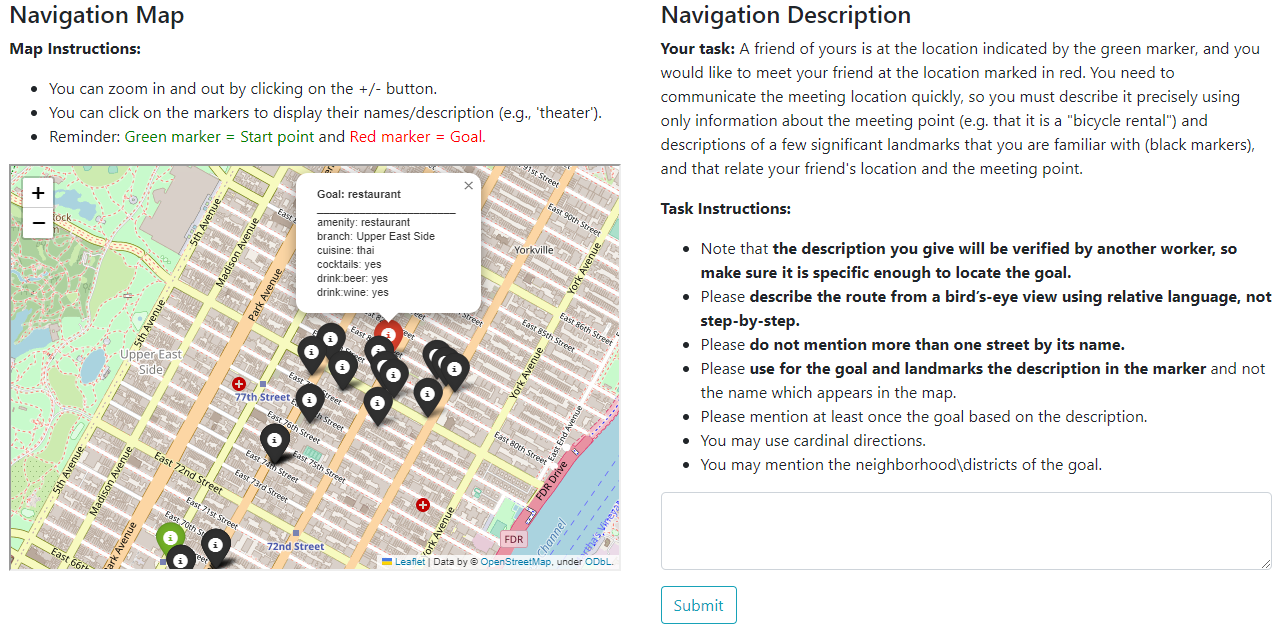}}
 \caption{Participant Interface: the instruction writing task.
 } 
\label{fig:instroctor}%

\end{figure*}

\begin{figure*}[h]
\centering
\scalebox{1}{
\includegraphics[width=\textwidth]{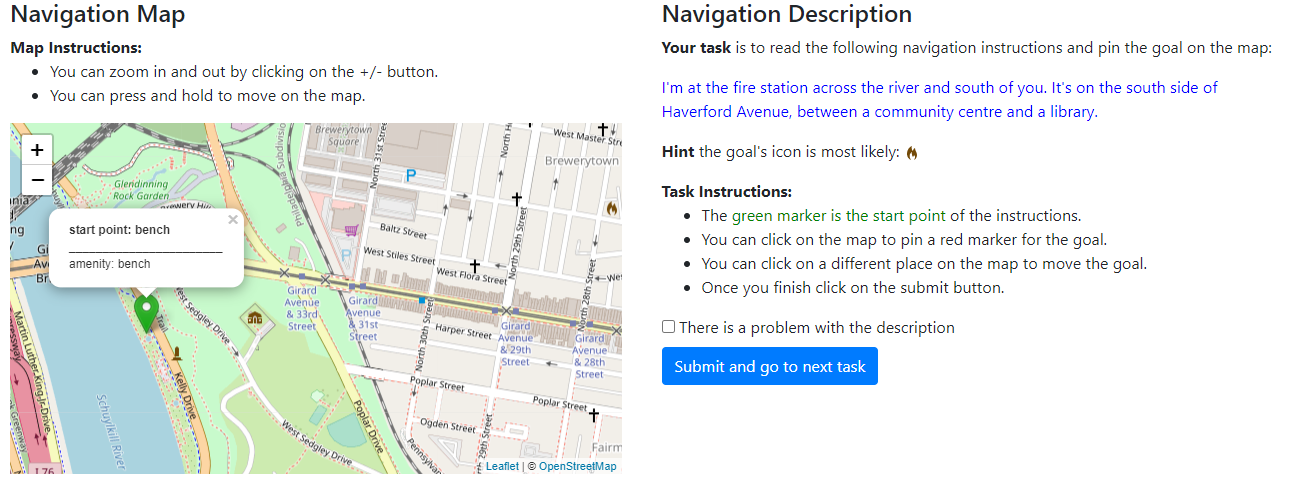}}
 \caption{Participant Interface: the validation task.
 } 
\label{fig:verify}%
\end{figure*}

\section{Results over Alternative Splits}
\label{app:results}

In Table \ref{tab:results} we showed the results on a split that was trained on Manhattan, with Pittsburgh as the development set and Philadelphia as the test set. However, Manhattan is demographically different from Pittsburgh and Philadelphia and contains more entities on the map. In Table \ref{tab:results_permutation} we show  results over different permutations of the cities -- testing on Manhattan and training on either Pittsburgh or Philadelphia. However, as the development Pittsburgh set and test Philadelphia sets contain few instructions (1,103 and 1,278 instructions, respectively), it seems they do not contain enough data to support learning. This claim is supported in Table \ref{tab:results_permutation} which shows the results for testing on Manhattan with different training sets. The T5 model, in both splits learns to predict close locations to the starting point, or even the exact location as the starting point. It therefore does not go over the limited range of 2K distance and has very low accuracy. The {\scshape T5+Graph} model has a higher accuracy but the model also predicts location over the limited range, resulting in a very high mean error distance. Additionally, the results for all models trained on Pittsburgh were slightly better than the ones trained on Philadelphia, which might be due to the size of the region, Philadelphia being more than twice as large as Pittsburgh, the  {\scshape T5+Graph} model struggles to learn connections --- i.e., grounding. --- between text and the environment.

\section{Participant Application Interface}
\label{app:UI}

The tasks are performed via an online assignment application, depicted in Figures \ref{fig:instroctor} and \ref{fig:verify}.

\end{document}